\documentclass{article}
\usepackage{amsfonts}
\usepackage{amsmath}
\usepackage{enumerate}
\usepackage{graphicx}
\usepackage{hyperref}
\usepackage{breakurl}
\usepackage[accepted]{icml2013}

\allowdisplaybreaks[4]
\let\originalleft\left
\let\originalright\right
\renewcommand{\left}{\mathopen{}\mathclose\bgroup\originalleft}
\renewcommand{\right}{\aftergroup\egroup\originalright}
\newcommand{\smallspace}{\;\;}
\newcommand{\largespace}{\;\;\;\;\;\;}
\DeclareMathOperator{\E}{E}
\DeclareMathOperator{\diag}{diag}
\newcommand{\mat}[1]{\mathbf{#1}}
\renewcommand{\vec}[1]{\boldsymbol{#1}}

\begin{document} 

\twocolumn[
\icmltitle{Kernelized Bayesian Matrix Factorization}

\icmlauthor{Mehmet G\"{o}nen $^{1,2}$}{mehmet.gonen@aalto.fi}
\icmlauthor{Suleiman A. Khan $^{1,2}$}{suleiman.khan@aalto.fi}
\icmlauthor{Samuel Kaski $^{1,2,3}$}{samuel.kaski@aalto.fi}
\icmladdress{$^{1}$ Helsinki Institute for Information Technology HIIT\\$^{2}$ Department of Information and Computer Science, Aalto University\\$^{3}$ Department of Computer Science, University of Helsinki}

\icmlkeywords{kernelized matrix factorization, variational approximation, biological interaction networks, multilabel learning}

\vskip 0.3in
]

%%%%%%%%%%%%%%%%%%%%%%%%%%%%%%%%%%%%%%%%%%%%%%%%%%%%%%%%%%%%%%%%%%%%%%%%%%%%%%%%%%%%%%%%%%%%%%%%%%%%

\begin{abstract}
We extend kernelized matrix factorization with a fully Bayesian treatment and with an ability to work with multiple side information sources expressed as different kernels. Kernel functions have been introduced to matrix factorization to integrate side information about the rows and columns (e.g., objects and users in recommender systems), which is necessary for making out-of-matrix (i.e., cold start) predictions. We discuss specifically bipartite graph inference, where the output matrix is binary, but extensions to more general matrices are straightforward. We extend the state of the art in two key aspects: (i)~A fully conjugate probabilistic formulation of the kernelized matrix factorization problem enables an efficient variational approximation, whereas fully Bayesian treatments are not computationally feasible in the earlier approaches. (ii)~Multiple side information sources are included, treated as different kernels in multiple kernel learning that additionally reveals which side information sources are informative. Our method outperforms alternatives in predicting drug--protein interactions on two data sets. We then show that our framework can also be used for solving multilabel learning problems by considering samples and labels as the two domains where matrix factorization operates on. Our algorithm obtains the lowest Hamming loss values on 10 out of 14 multilabel classification data sets compared to five state-of-the-art multilabel learning algorithms.
\end{abstract}

%%%%%%%%%%%%%%%%%%%%%%%%%%%%%%%%%%%%%%%%%%%%%%%%%%%%%%%%%%%%%%%%%%%%%%%%%%%%%%%%%%%%%%%%%%%%%%%%%%%%

\section{Introduction}
Matrix factorization algorithms are very popular matrix completion methods \citep{srebro04thesis}, successfully used in many applications including recommender systems and image inpainting. The main idea behind these methods is to factorize a partially observed data matrix by finding a low-dimensional latent representation for both its rows and columns. The prediction for a missing entry can be calculated as the inner product between the latent representations of the corresponding row and column. \citet{salakhutdinov08anips,salakhutdinov08bicml} give a probabilistic formulation for matrix factorization and its fully Bayesian extension. However, these approaches are still incomplete in two major aspects: (i)~It is not possible to integrate side information (e.g., social network or user profiles for recommender systems) into the model. (ii)~It is not possible to make predictions for completely empty columns or rows (i.e., out-of-matrix prediction).

Algorithms for integrating side information into matrix factorization have been proposed earlier in the recommender systems literature. \citet{ma08cikm} propose a probabilistic matrix factorization method that uses a social network and the rating matrix together to find better latent components. \citet{shan10icdm} integrate side information into a probabilistic matrix factorization model using topic models to generate latent components of the rated items (e.g., movies). \citet{agarwal10wsdm} use a similar strategy to generate latent components of both users and items using topic models. \citet{wang11kdd} also combine matrix factorization and topic models for scientific article recommendation using textual content of articles as side information. All these algorithms are based on explicit feature representations; some are specific to count (e.g., text) data, and all are able to model linear dependencies. We use kernels to include nonlinear dependencies.

\citet{lawrence09icml} formulate a nonlinear matrix factorization method by generating latent components via Gaussian processes without integrating any side information. Recently, \citet{zhou12sdm} propose a kernelized probabilistic matrix factorization method using Gaussian process priors with covariance matrices on side information. However, with the modeling assumptions, only a {\it maximum a posteriori} (MAP) estimate for the latent components is computationally feasible, and even then the used gradient descent approach can be very slow. Furthermore, the method uses only a single kernel for each domain and needs test instances while training to be able to calculate their latent components (i.e., {\it transductive learning}).

In this paper, we focus on modeling interaction networks between two domains (e.g., biological networks between drugs and proteins), and estimating unknown interactions between objects from these two domains, which is also known as {\it bipartite graph inference} \citep{yamanishi09nips}. The standard pairwise kernel approaches are based on a kernel matrix over object pairs in the training set and are computationally expensive \citep{benhur05bio}. There are also kernel-based (non-Bayesian) dimensionality reduction algorithms that map objects from both domains into the same subspace and perform prediction there \citep{yamanishi09nips, yamanishi08bio, yamanishi10bio}.

In biological interaction networks, being composed of structured objects such as drugs and proteins, there exist several feature representations or similarity measures for the objects \citep{scholkopf04book}. Instead of using a single specific kernel, we can combine multiple kernel functions to obtain a better similarity measure, which is known as {\it multiple kernel learning} \citep{gonen11jmlr}. 

We introduce a kernelized Bayesian matrix factorization method and give its details for the bipartite graph inference scenario; it can also be applied to other types of matrices with slight modifications. Our two main contributions are: (i)~We formulate a novel fully conjugate probabilistic model that allows us to develop an efficient variational approximation scheme, the first fully Bayesian treatment which is still significantly faster than the earlier method for computing MAP point estimates \citep{zhou12sdm}. (ii)~The proposed method is able to integrate multiple side information sources by coupling matrix factorization with multiple kernel learning. We show the effectiveness of our approach on one toy data set and two drug--protein interaction data sets. We then show how our method can be used to solve multilabel learning problems and report classification results on 14 benchmark data sets.

%%%%%%%%%%%%%%%%%%%%%%%%%%%%%%%%%%%%%%%%%%%%%%%%%%%%%%%%%%%%%%%%%%%%%%%%%%%%%%%%%%%%%%%%%%%%%%%%%%%%

\section{Preliminaries and Notation} \label{sec:notation}
We assume that the objects come from two domains $\mathcal{X}$ and $\mathcal{Z}$. We are given two samples of independent and identically distributed training instances from each, denoted by $\mat{X} = \{\vec{x}_{i} \in \mathcal{X}\}_{i = 1}^{N_{\texttt{x}}}$ and $\mat{Z} = \{\vec{z}_{j} \in \mathcal{Z}\}_{j = 1}^{N_{\texttt{z}}}$. In order to calculate similarities between the objects of the same domain, we have multiple kernel functions for each domain, namely, $\{k_{\texttt{x},m}\colon \mathcal{X} \times \mathcal{X} \to \mathbb{R}\}_{m = 1}^{P_{\texttt{x}}}$ and $\{k_{\texttt{z},n}\colon \mathcal{Z} \times \mathcal{Z} \to \mathbb{R}\}_{n = 1}^{P_{\texttt{z}}}$. If the side information comes in the form of features instead of similarities, the set of kernels defined for a specific domain correspond to different notions of similarity on the same feature representation or may be using information coming from multiple feature representations (i.e., views).

The $(i,j)$th entry of the target label matrix $\mat{Y} \in \{-1,+1\}^{N_{\texttt{x}} \times N_{\texttt{z}}}$ is
\begin{align*}
	y_{j}^{i} = \begin{cases}
					+1 & \textnormal{if $\vec{x}_{i}$ and $\vec{z}_{j}$ are interacting,} \\
					-1 & \textnormal{otherwise.}
				\end{cases}
\end{align*}
The superscript indexes the rows and the subscript indexes the columns. The prediction task is to estimate unknown interactions for out-of-matrix objects, which is also known as {\it cold start} prediction in recommender systems.

\begin{figure}[!htb]
	\centering	
	\includegraphics[scale=0.80]{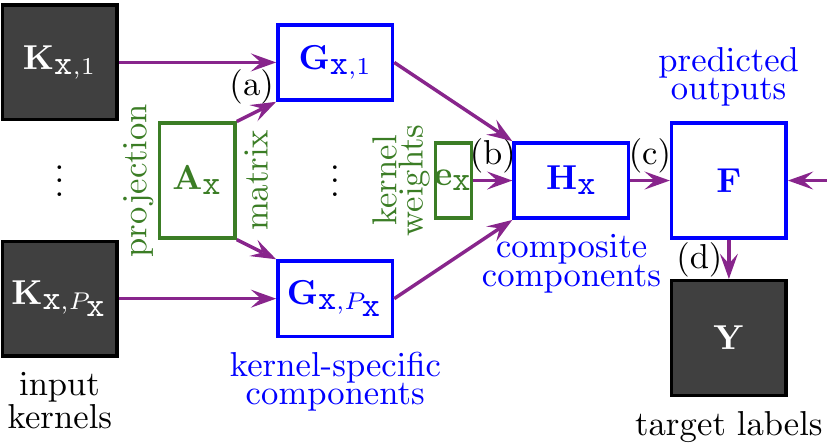}
	\vskip -0.1in
	\caption{Flowchart of kernelized matrix factorization with twin multiple kernel learning (the twin domain $\mathcal{Z}$ is omitted for clarity).} \label{fig:kbmf2mkl_flow}
\end{figure}

Figure~\ref{fig:kbmf2mkl_flow} illustrates the method we propose; it is composed of four main parts: (a) kernel-based nonlinear dimensionality reduction, (b) multiple kernel learning, (c) matrix factorization, and (d) binary classification. The first two kernel-based parts are applied to each domain separately and they are completely symmetric, hence we call them {\it twins}. One of the twins (i.e., the one that operates on domain $\mathcal{Z}$) is omitted for clarity. In this section, we briefly explain each part and introduce the notation used. In the following sections, we formulate a fully conjugate probabilistic model and derive a variational approximation.

{\bf Kernel-Based Nonlinear Dimensionality Reduction.} In this part, we perform feature extraction using the input kernel matrices $\{\mat{K}_{\texttt{x},m} \in \mathbb{R}^{N_{\texttt{x}} \times N_{\texttt{x}}}\}_{m = 1}^{P_{\texttt{x}}}$ and the common projection matrix $\mat{A}_{\texttt{x}} \in \mathbb{R}^{N_{\texttt{x}} \times R}$ where $R$ is the corresponding subspace dimensionality. We obtain the kernel-specific components $\{\mat{G}_{\texttt{x}, m} = \mat{A}_{\texttt{x}}^{\top} \mat{K}_{\texttt{x},m}\}_{m = 1}^{P_{\texttt{x}}}$ after the projection. The main idea is very similar to {\it kernel principal component analysis} or {\it kernel Fisher discriminant analysis}, where the columns of the projection matrix can be solved with eigendecompositions \citep{scholkopf02book}. However, this solution strategy is not possible for the more complex model formulated here.

{\bf Multiple Kernel Learning.} This part is responsible for combining the kernel-specific components linearly to obtain the composite components $\mat{H}_{\texttt{x}} = \sum \nolimits_{m = 1}^{P_{\texttt{x}}} e_{\texttt{x},m} \mat{G}_{\texttt{x},m}$ where the kernel weights can take arbitrary values $\vec{e}_{\texttt{x}} \in \mathbb{R}^{P_{\texttt{x}}}$. If we have a single kernel function for a specific domain, we can safely ignore the composite components and the kernel weights, and use the single available kernel-specific components to represent the objects in that domain \citep{gonen12bio}. The details of our method with a single kernel function for each domain are explained in the supplementary material.

{\bf Matrix Factorization.} In this part, we propose to use the low-dimensional representations of objects in the unified subspace, namely, $\mat{H}_{\texttt{x}}$ and $\mat{H}_{\texttt{z}}$, to calculate the predicted output matrix $\mat{F} = \mat{H}_{\texttt{x}}^{\top} \mat{H}_{\texttt{z}}$. This corresponds to factorizing the predicted outputs into two low-rank matrices.

{\bf Binary Classification.} This part just assigns a class label to each object pair $(\vec{x}_{i}, \vec{z}_{j})$ by looking at the sign of the predicted output $f_{j}^{i}$ in the matrix factorization part. The proposed method can also be extended to handle other types of outputs (e.g., real-valued outputs used in recommender systems) by removing the binary classification part and directly generating the target outputs in the matrix factorization part. This corresponds to removing the predicted output matrix $\mat{F}$ and generating target label matrix $\mat{Y}$ directly from the composite components $\mat{H}_{\texttt{x}}$ and $\mat{H}_{\texttt{z}}$. The details of our method for real-valued outputs are also given in the supplementary material.

%%%%%%%%%%%%%%%%%%%%%%%%%%%%%%%%%%%%%%%%%%%%%%%%%%%%%%%%%%%%%%%%%%%%%%%%%%%%%%%%%%%%%%%%%%%%%%%%%%%%

\section{Kernelized Bayesian Matrix Factorization with Twin Multiple Kernel Learning} \label{sec:kbmf2mkl}
For the method described in the previous section, we formulate a probabilistic model, called {\it kernelized Bayesian matrix factorization with twin multiple kernel learning} (KBMF2MKL), which has two key properties that enable us to perform efficient inference: (i)~The kernel-specific and composite components are modeled explicitly by introducing them as latent variables. (ii)~Kernel weights are assumed to be normally distributed without enforcing any constraints (e.g., non-negativity) on them. The reasons for introducing these two properties to our probabilistic model becomes clear when we explain our inference method.

\begin{figure}[!htb]
	\centering
	\includegraphics[scale=0.80]{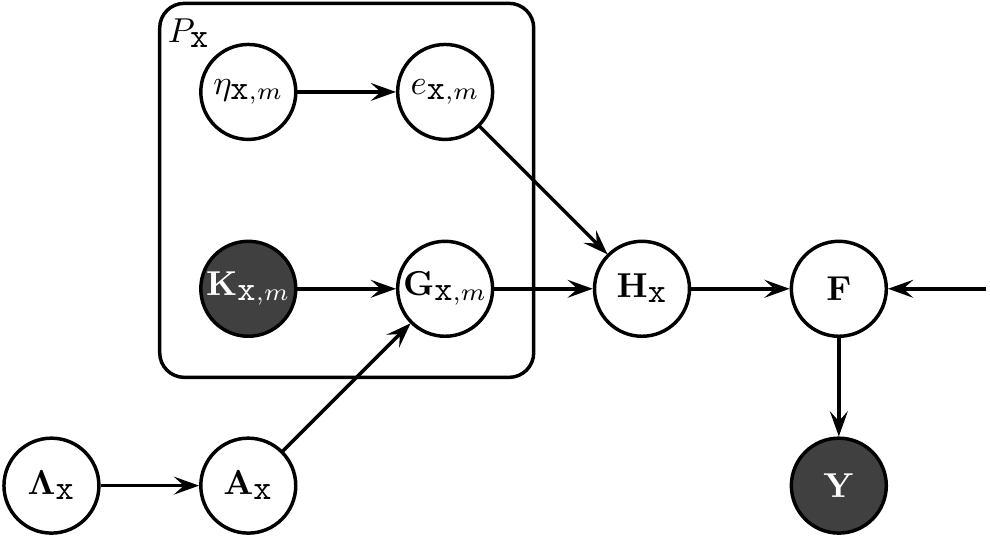}
	\vskip -0.1in
	\caption{Graphical model of kernelized Bayesian matrix factorization with twin multiple kernel learning.} \label{fig:kbmf2mkl_graphical}
\end{figure}

Figure~\ref{fig:kbmf2mkl_graphical} gives the graphical model of KBMF2MKL with latent variables and their corresponding priors. There are some additions to the notation described earlier: The $N_{\texttt{x}} \times R$ matrix of priors for the entries of the projection matrix $\mat{A}_{\texttt{x}}$ is denoted by $\mat{\Lambda}_{\texttt{x}}$. The $P_{\texttt{x}} \times 1$ vector of priors for the kernel weights $\vec{e}_{\texttt{x}}$ is denoted by $\vec{\eta}_{\texttt{x}}$. The standard deviations for the kernel-specific and composite components are represented as $\sigma_{g}$ and $\sigma_{h}$, respectively; these hyper-parameters are not shown for clarity.

The distributional assumptions of the dimensionality reduction part are
\begin{alignat*}{4}
	\lambda_{\texttt{x},s}^{i} &\sim \mathcal{G}(\lambda_{\texttt{x},s}^{i}; \alpha_{\lambda}, \beta_{\lambda}) &&\largespace \forall (i, s) \\
	a_{\texttt{x},s}^{i} | \lambda_{\texttt{x},s}^{i} &\sim \mathcal{N}(a_{\texttt{x},s}^{i}; 0, (\lambda_{\texttt{x},s}^{i})^{-1}) &&\largespace \forall (i, s) \\
	g_{\texttt{x},m,i}^{s} | \vec{a}_{\texttt{x},s}, \vec{k}_{\texttt{x},m,i} &\sim \mathcal{N}(g_{\texttt{x},m,i}^{s}; \vec{a}_{\texttt{x},s}^{\top} \vec{k}_{\texttt{x},m,i}, \sigma_{g}^{2}) &&\largespace \forall (m, s, i)
\end{alignat*}
where $\mathcal{N}(\cdot; \vec{\mu}, \mat{\Sigma})$ is the normal distribution with mean vector $\vec{\mu}$ and covariance matrix $\mat{\Sigma}$, and $\mathcal{G}(\cdot; \alpha, \beta)$ denotes the gamma distribution with shape parameter $\alpha$ and scale parameter $\beta$. The multiple kernel learning part has the following distributional assumptions:
\begin{alignat*}{4}
	\eta_{\texttt{x},m} &\sim \mathcal{G}(\eta_{\texttt{x},m}; \alpha_{\eta}, \beta_{\eta}) &&\smallspace \forall m \\\
	e_{\texttt{x},m} | \eta_{\texttt{x},m} &\sim \mathcal{N}(e_{\texttt{x},m}; 0, \eta_{\texttt{x},m}^{-1}) &&\smallspace \forall m \\
	h_{\texttt{x},i}^{s} | \{e_{\texttt{x}, m}, g_{\texttt{x},m,i}^{s}\}_{m = 1}^{P_{\texttt{x}}} &\sim \mathcal{N}\left(h_{\texttt{x},i}^{s}; \sum \limits_{m = 1}^{P_{\texttt{x}}} e_{\texttt{x},m} g_{\texttt{x},m,i}^{s}, \sigma_{h}^{2}\right) &&\smallspace \forall (s, i)
\end{alignat*}
where kernel-level sparsity can be tuned by changing the hyper-parameters $(\alpha_{\eta}, \beta_{\eta})$. Setting the gamma priors to induce sparsity, e.g., $(\alpha_{\eta}, \beta_{\eta}) = (0.001, 1000)$, produces results analogous to using the $\ell_{1}$-norm on the kernel weights, whereas using uninformative priors, e.g., $(\alpha_{\eta}, \beta_{\eta}) = (1, 1)$, resembles using the $\ell_{2}$-norm. The matrix factorization part calculates the predicted outputs using the inner products between the low-dimensional representations of the object pairs:
\begin{align*}
	f_{j}^{i} | \vec{h}_{\texttt{x},i}, \vec{h}_{\texttt{z},j} \sim \mathcal{N}(f_{j}^{i}; \vec{h}_{\texttt{x},i}^{\top} \vec{h}_{\texttt{z},j}, 1) \largespace \forall (i, j)
\end{align*}
where the predicted outputs are introduced to speed up the inference procedures \citep{albert93jasa}. The binary classification part forces the predicted outputs to have the same sign with their target labels:
\begin{align*}
	y_{j}^{i} | f_{j}^{i} \sim \delta(f_{j}^{i} y_{j}^{i} > \nu) \largespace \forall (i, j)
\end{align*}
where the margin parameter $\nu$ is introduced to remove ambiguity in the scaling and to place a low-density region between two classes, similar to the margin idea in SVMs, which is generally used for semi-supervised learning \citep{lawrence05nips}. Here, $\delta(\cdot)$ is the Kronecker delta function that returns 1 if its argument is true and 0 otherwise.

%%%%%%%%%%%%%%%%%%%%%%%%%%%%%%%%%%%%%%%%%%%%%%%%%%%%%%%%%%%%%%%%%%%%%%%%%%%%%%%%%%%%%%%%%%%%%%%%%%%%

\section{Efficient Inference Using Variational Approximation} \label{sec:inference}
Exact inference for the model is intractable and of the two readily available alternatives, Gibbs sampling and variational approximation, we choose the latter for computational efficiency. Variational methods optimize a lower bound on the marginal likelihood, which involves a factorized approximation of the posterior, to find the joint parameter distribution \citep{beal03thesis}. 

As short-hand notations, all hyper-parameters in the model are denoted by $\vec{\zeta} = \{\alpha_{\eta}, \beta_{\eta}, \alpha_{\lambda}, \beta_{\lambda}, \sigma_{g}, \sigma_{h}, \nu\}$, all prior variables by $\mat{\Xi} = \{\vec{\eta}_{\texttt{x}}, \vec{\eta}_{\texttt{z}}, \mat{\Lambda}_{\texttt{x}}, \mat{\Lambda}_{\texttt{z}}\}$, and the remaining random variables by $\mat{\Theta} = \{\mat{A}_{\texttt{x}}, \mat{A}_{\texttt{z}}, \vec{e}_{\texttt{x}}, \vec{e}_{\texttt{z}}, \mat{F}, \{\mat{G}_{\texttt{x},m}\}_{m = 1}^{P_{\texttt{x}}}, \{\mat{G}_{\texttt{z},n}\}_{n = 1}^{P_{\texttt{z}}}, \mat{H}_{\texttt{x}}, \mat{H}_{\texttt{z}}\}$. We omit the dependence on $\vec{\zeta}$ for clarity. We factorize the variational approximation as
\begin{multline*}
	p(\mat{\Theta}, \mat{\Xi} | \{\mat{K}_{\texttt{x},m}\}_{m = 1}^{P_{\texttt{x}}}, \{\mat{K}_{\texttt{z},n}\}_{n = 1}^{P_{\texttt{z}}}, \mat{Y}) \approx q(\mat{\Theta}, \mat{\Xi}) = \\ q(\mat{\Lambda}_{\texttt{x}}) q(\mat{A}_{\texttt{x}}) q(\{\mat{G}_{\texttt{x},m}\}_{m = 1}^{P_{\texttt{x}}}) q(\vec{\eta}_{\texttt{x}}) q(\vec{e}_{\texttt{x}}) q(\mat{H}_{\texttt{x}})\\ q(\mat{\Lambda}_{\texttt{z}}) q(\mat{A}_{\texttt{z}}) q(\{\mat{G}_{\texttt{z},n}\}_{n = 1}^{P_{\texttt{z}}}) q(\vec{\eta}_{\texttt{z}}) q(\vec{e}_{\texttt{z}}) q(\mat{H}_{\texttt{z}}) q(\mat{F})
\end{multline*}
and define each factor according to its full conditional:
\begin{align*}
	q(\mat{\Lambda}_{\texttt{x}}) &= \prod \limits_{i = 1}^{N_{\texttt{x}}} \prod \limits_{s = 1}^{R} \mathcal{G}(\lambda_{\texttt{x},s}^{i}; \alpha(\lambda_{\texttt{x},s}^{i}), \beta(\lambda_{\texttt{x},s}^{i}))\\
	q(\mat{A}_{\texttt{x}}) &= \prod \limits_{s = 1}^{R} \mathcal{N}(\vec{a}_{\texttt{x},s}; \mu(\vec{a}_{\texttt{x},s}), \Sigma(\vec{a}_{\texttt{x},s})) \\
	q(\{\mat{G}_{\texttt{x},m}\}_{m = 1}^{P_{\texttt{x}}}) &= \prod \limits_{m = 1}^{P_{\texttt{x}}} \prod \limits_{i = 1}^{N_{\texttt{x}}} \mathcal{N}(\vec{g}_{\texttt{x},m,i}; \mu(\vec{g}_{\texttt{x},m,i}), \Sigma(\vec{g}_{\texttt{x},m,i})) \\
		q(\vec{\eta}_{\texttt{x}}) &= \prod \limits_{m = 1}^{P_{\texttt{x}}} \mathcal{G}(\eta_{\texttt{x},m}; \alpha(\eta_{\texttt{x},m}), \beta(\eta_{\texttt{x},m})) \\
	q(\vec{e}_{\texttt{x}}) &= \mathcal{N}(\vec{e}_{\texttt{x}}; \mu(\vec{e}_{\texttt{x}}), \Sigma(\vec{e}_{\texttt{x}})) \\
	q(\mat{H}_{\texttt{x}}) &= \prod \limits_{i = 1}^{N_{\texttt{x}}} \mathcal{N}(\vec{h}_{\texttt{x},i}; \mu(\vec{h}_{\texttt{x},i}), \Sigma(\vec{h}_{\texttt{x},i})) \\
	q(\mat{F}) &= \prod \limits_{i = 1}^{N_{\texttt{x}}} \prod \limits_{j = 1}^{N_{\texttt{z}}} \mathcal{TN}(f_{j}^{i}; \mu(f_{j}^{i}), \Sigma(f_{j}^{i}), \rho(f_{j}^{i}))
\end{align*}
where $\alpha(\cdot)$, $\beta(\cdot)$, $\mu(\cdot)$, and $\Sigma(\cdot)$ denote the shape parameter, scale parameter, mean vector, and covariance matrix, respectively. Here, $\mathcal{TN}(\cdot; \vec{\mu}, \mat{\Sigma}, \rho(\cdot))$ denotes the truncated normal distribution with mean vector $\vec{\mu}$, covariance matrix $\mat{\Sigma}$, and truncation rule $\rho(\cdot)$ such that $\mathcal{TN}(\cdot; \vec{\mu}, \mat{\Sigma}, \rho(\cdot)) \propto \mathcal{N}(\cdot; \vec{\mu}, \mat{\Sigma})$ if $\rho(\cdot)$ is true and $\mathcal{TN}(\cdot; \vec{\mu}, \mat{\Sigma}, \rho(\cdot)) = 0$ otherwise.

We can bound the marginal likelihood using Jensen's inequality:
\begin{multline*}
	\log p(\mat{Y} | \{\mat{K}_{\texttt{x},m}\}_{m = 1}^{P_{\texttt{x}}}, \{\mat{K}_{\texttt{z},n}\}_{n = 1}^{P_{\texttt{z}}}) \geq \\ \E_{q(\mat{\Theta}, \mat{\Xi})}[\log p(\mat{Y}, \mat{\Theta}, \mat{\Xi} | \{\mat{K}_{\texttt{x},m}\}_{m = 1}^{P_{\texttt{x}}}, \{\mat{K}_{\texttt{z},n}\}_{n = 1}^{P_{\texttt{z}}})] \\- \E_{q(\mat{\Theta}, \mat{\Xi})}[\log q(\mat{\Theta}, \mat{\Xi})]
\end{multline*}
and optimize this bound by maximizing with respect to each factor separately until convergence. The approximate posterior distribution of a specific factor $\vec{\tau}$ can be found as
\begin{multline*}
	q(\vec{\tau}) \propto \\ \exp(\E_{q(\{\mat{\Theta}, \mat{\Xi}\} \backslash \vec{\tau} )}[\log p(\mat{Y}, \mat{\Theta}, \mat{\Xi} | \{\mat{K}_{\texttt{x},m}\}_{m = 1}^{P_{\texttt{x}}}, \{\mat{K}_{\texttt{z},n}\}_{n = 1}^{P_{\texttt{z}}})]).
\end{multline*}
For our model, thanks to the conjugacy, the resulting approximate posterior distribution of each factor follows the same distribution as the corresponding factor. The variational updates for the approximate posterior distributions are given in the supplementary material.

{\bf Modeling Choices.} Note that using the kernel-specific and composite components as latent variables in our probabilistic model introduces extra conditional independencies between the random variables and enables us to derive efficient update rules for the approximate posterior distributions. The other key property of our model is the assumption of normality of the kernel weights, which allows us to obtain a fully conjugate probabilistic model \citep{gonen12icml}. In earlier Bayesian multiple kernel learning algorithms, the combined kernel has usually been defined as a convex sum of the input kernels, by assuming a Dirichlet distribution on the kernel weights \citep{girolami05icml,damoulas08bio}. As a consequence of the nonconjugacy between Dirichlet and normal distributions, they need to use a sampling strategy (e.g., importance sampling) to update the kernel weights when deriving variational approximations.

{\bf Convergence.} The inference mechanism sequentially updates the approximate posterior distributions of the latent variables and the corresponding priors until convergence, which can be checked by monitoring the lower bound. The first term of the lower bound corresponds to the sum of exponential-form expectations of the distributions in the joint likelihood. The second term is the sum of negative entropies of the approximate posteriors in the ensemble. The only nonstandard distribution in these terms is the truncated normal distribution used for the predicted outputs, and the truncated normal distribution has a closed-form formula also for its entropy.

{\bf Computational Complexity.} The most time-consuming operations of the update equations are covariance calculations because they need matrix inversions. The time complexity of the covariance updates for the projection matrices is $\mathcal{O}(R \max(N_{\texttt{x}}^{3}, N_{\texttt{z}}^{3}))$ and we can cache $\sum \nolimits_{m = 1}^{P_\texttt{x}} \mat{K}_{\texttt{x},m} \mat{K}_{\texttt{x},m}^{\top}$ and $\sum \nolimits_{n = 1}^{P_\texttt{z}} \mat{K}_{\texttt{z},n} \mat{K}_{\texttt{z},n}^{\top}$ before starting inference procedure to reduce the computational cost of these updates. The covariance updates for the kernel-specific components require inverting diagonal matrices. The time complexities of the covariance updates for the kernel weights and the composite components are $\mathcal{O}(\max(P_{\texttt{x}}^{3}, P_{\texttt{z}}^{3}))$. The other calculations in these updates can be done efficiently using matrix-matrix or matrix-vector multiplications. Finding the posterior expectations of the predicted outputs only requires evaluating the standardized normal cumulative distribution function and the standardized normal probability density. In summary, the total time complexity of each iteration in our variational approximation scheme is $\mathcal{O}(R \max(N_{\texttt{x}}^{3}, N_{\texttt{z}}^{3}) + \max(P_{\texttt{x}}^{3}, P_{\texttt{z}}^{3}))$, which makes the algorithm more efficient than standard pairwise kernel approaches \citep{benhur05bio} that require calculating a kernel matrix over pairs and training a kernel-based classifier using this kernel, resulting in $\mathcal{O}(N_{\texttt{x}}^{3} N_{\texttt{z}}^{3})$ complexity.

{\bf Prediction.} Given a test pair $(\vec{x}_{\star}, \vec{z}_{*})$, we want to predict the corresponding score $f_{*}^{\star}$ or target label $y_{*}^{\star}$. We first replace posterior distributions of $\mat{A}_{\texttt{x}}$, $\mat{A}_{\texttt{z}}$, $\vec{e}_{\texttt{x}}$, and $\vec{e}_{\texttt{z}}$ with their approximate posterior distributions $q(\mat{A}_{\texttt{x}})$, $q(\mat{A}_{\texttt{z}})$, $q(\vec{e}_{\texttt{x}})$, and $q(\vec{e}_{\texttt{z}})$. Using the approximate distributions, we obtain the predictive distributions of the kernel-specific and composite components. The predictive distribution of the target label can finally be formulated as
\begin{multline*}
	p(y_{*}^{\star} = +1 | \{\vec{k}_{\texttt{x},m,\star}, \mat{K}_{\texttt{x},m}\}_{m = 1}^{P_\texttt{x}}, \{\vec{k}_{\texttt{z},n,*}, \mat{K}_{\texttt{z},n}\}_{n = 1}^{P_\texttt{z}}, \mat{Y}) = \\(\mathcal{Z}_{*}^{\star})^{-1} \Phi\left(\dfrac{\mu(f_{*}^{\star}) - \nu}{\Sigma(f_{*}^{\star})}\right)
\end{multline*}
where $\mathcal{Z}_{*}^{\star}$ is the normalization coefficient calculated for the test pair and $\Phi(\cdot)$ is the standardized normal cumulative distribution function.

%%%%%%%%%%%%%%%%%%%%%%%%%%%%%%%%%%%%%%%%%%%%%%%%%%%%%%%%%%%%%%%%%%%%%%%%%%%%%%%%%%%%%%%%%%%%%%%%%%%%

\section{Experiments} \label{sec:experiments}
We first run our method on a toy data set to illustrate its kernel learning capability. We then test its performance in a real-life application with experiments on two drug--protein interaction data sets. One of them is a standard data set with a single view for each domain and the other one is a larger multiview data set we have collected. We also perform experiments on 14 benchmark multilabel classification data sets in order to show the suitability of our matrix factorization framework with side information in a nonstandard application scenario. Our Matlab implementations are available at \burl{http://research.ics.aalto.fi/mi/software/kbmf/}. 

\subsection{Toy Data Set}
We create a toy data set consisting of samples from two domains and real-valued outputs for object pairs. The data generation process is:
\begin{alignat*}{4}
	x_{i}^{m} &\sim \mathcal{N}(x_{i}^{m}; 0, 1) &&\largespace \forall (m, i) \\
	z_{j}^{n} &\sim \mathcal{N}(z_{j}^{n}; 0, 1) &&\largespace \forall (n, j) \\
	y_{j}^{i} | \vec{x}_{i}, \vec{z}_{j} &\sim \mathcal{N}(y_{j}^{i}; x_{i}^{1} z_{j}^{3} + x_{i}^{4} z_{j}^{8} + x_{i}^{7} z_{j}^{10}, 1) &&\largespace \forall (i, j)
\end{alignat*}
where $(N_{\texttt{x}}, N_{\texttt{z}}) = (40, 60)$, the samples from $\mathcal{X}$ and $\mathcal{Z}$ are generated from 15- and 10-dimensional isotropic normal distributions with unit variance (i.e., $m \in \{1,\dots,15\}$ and $n \in \{1,\dots,10\}$), respectively, and the target outputs are generated using only three features from each domain. Note that this data set has not been generated from our probabilistic model.

In order to have multiple kernel functions for each domain, we calculate a separate linear kernel for each feature of the data points, i.e., $(P_{\texttt{x}}, P_{\texttt{z}}) = (15, 10)$. We then learn our model, intended to work as a predictive model that identifies the relevant features for the prediction task and has a good generalization performance. We use uninformative priors for the projection matrices and the kernel weights by setting $(\alpha_{\eta}, \beta_{\eta}, \alpha_{\lambda}, \beta_{\lambda}) = (1, 1, 1, 1)$. The standard deviations are set to $(\sigma_{g}, \sigma_{h}, \sigma_{y}) = (0.1, 0.1, 1)$, where $\sigma_{y}$ denotes the noise level used for the target outputs. The subspace dimensionality is arbitrarily set to five (i.e., $R = 5$).

\begin{figure}[!htb]
	\centering
	\includegraphics[scale=0.40]{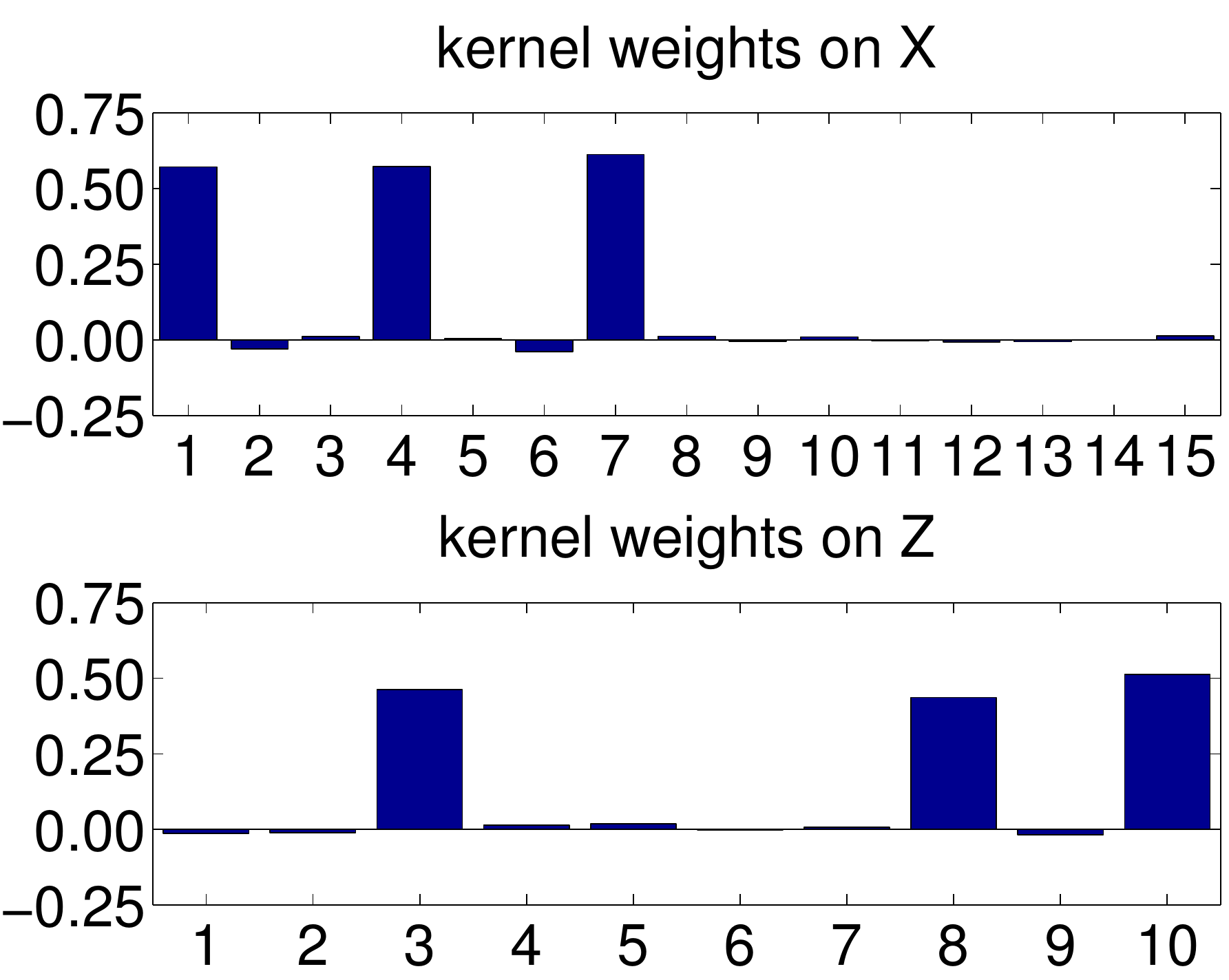}
	\vskip -0.1in
	\caption{Posterior means of the kernel weights found by our method on the toy data set.} \label{fig:toy_results}
\end{figure}

Figure~\ref{fig:toy_results} shows the found posterior means of the kernel weights. We see that our method correctly identifies the relevant features for each domain (i.e., the first, fourth, and seventh features for $\mathcal{X}$ and the third, eighth, and tenth features for $\mathcal{Z}$). The {\it root mean square error} between the target and predicted outputs is 0.9865 in accordance with the level of noise added.

\subsection{Drug--Protein Interaction Data Sets}
As the first case study, we analyze a drug--protein interaction network of humans, involving enzymes in particular. This drug--protein interaction network contains 445 drugs, 664 proteins, and 2926 experimentally validated interactions between them. The data set consists of the chemical similarity matrix between drugs, the genomic similarity matrix between proteins, and the target matrix of known interactions provided by \citet{yamanishi08bio}.

We compare one baseline and three matrix factorization methods: (i)~\texttt{Baseline} simply calculates the target output averages over each column or row as the predictions, (ii)~{\it kernelized probabilistic matrix factorization} (\texttt{KPMF}) method of \citet{zhou12sdm} with real-valued outputs, (iii)~our {\it kernelized Bayesian matrix factorization} (\texttt{KBMF}) method with real-valued outputs, and (iv)~\texttt{KBMF} method with binary outputs.

Our experimental methodology is as follows: For \texttt{KPMF}, the standard deviation $\sigma_{y}$ is set to one. For both \texttt{KBMF} variants, we use uninformative priors for the projection matrices and the kernel weights, i.e., $(\alpha_{\eta}, \beta_{\eta}, \alpha_{\lambda}, \beta_{\lambda}) = (1, 1, 1, 1)$, and the standard deviations $(\sigma_{g}, \sigma_{h})$ are set to $(0.1, 0.1)$. For \texttt{KBMF} with real-valued outputs, the standard deviation $\sigma_{y}$ is set to one. For \texttt{KBMF} with binary outputs, the margin parameter $\nu$ is arbitrarily set to one. We perform simulations with eight different numbers of components, i.e., $R \in \{5, 10, \dots, 40\}$. We run five replications of five-fold cross validation over drugs and report the average {\it area under ROC curve} (AUC) over the 25 results as the performance measure.

In the results, \texttt{C} and \texttt{G} mark the chemical similarity between drugs and the genomic similarity between proteins, respectively, whereas \texttt{N} marks the similarity between proteins calculated from the interaction network and it is defined as the ratio between (i)~the number of drugs that are interacting with both proteins and (ii)~the number of drugs that are interacting with at least one of the proteins, (i.e., Jaccard index). 

\begin{figure}[!htb]
	\centering
	\includegraphics[scale=0.40]{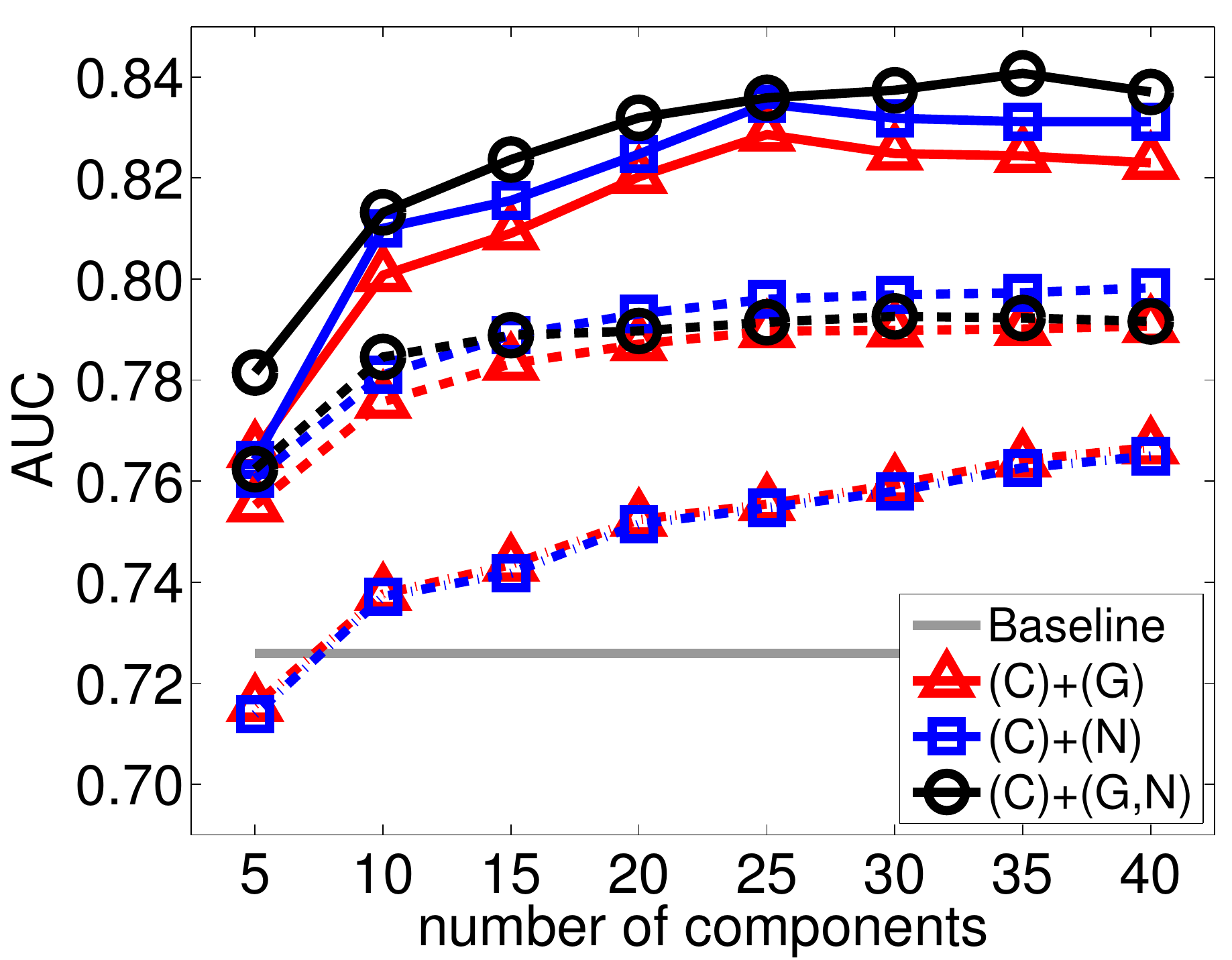}
	\vskip -0.1in
	\caption{Average prediction performances (area under ROC curve, AUC) on the drug--protein data set of \citet{yamanishi08bio}. Gray solid line: \texttt{Baseline}; other solid lines: \texttt{KBMF} with binary outputs; dashed lines: \texttt{KBMF}; dash-dotted lines: \texttt{KPMF}.} \label{fig:enzyme_results}
\end{figure}

The results in Figure~\ref{fig:enzyme_results} reveal that \texttt{KPMF} is above the baseline for more than 5 components, and both variants of \texttt{KBMF} for all component numbers. Both variants of our new \texttt{KBMF} outperform the earlier \texttt{KPMF} for all types of inputs, where the differences between \texttt{KPMF} and \texttt{KBMF} are statistically significant (paired $t$-test, $p < 0.01$). The difference is not due to \texttt{KPMF} having been introduced only for real-valued outputs, as even the real-output variant of \texttt{KBMF} is better. The difference is not due to the inability of the current version of \texttt{KPMF} to handle multiple data views either, as the single-kernel \texttt{KBMF} outperforms it. Hence the differences in the performance are due to the differences in the inference: MAP point estimates versus fully Bayesian inference. The best results are obtained with the binary-output \texttt{KBMF} when using all data sources.

Note that when we combine the genomic and network similarities between proteins using our method, the resulting similarity measure for proteins is better than those of single-kernel scenarios, leading to better prediction performance. This shows that when we have multiple side information sources about the objects, integrating them into the matrix factorization model in a principled way improves the results.

We study an additional drug--protein interaction network of humans, containing 855 drugs, 800 proteins, and 4659 experimentally validated interactions between them, extracted from the drugs and proteins of the data set provided by \citet{khan12bmcbio}. We have two views consisting of two standard 3D chemical structure descriptors for drugs, namely, 1120-dimensional {\it Amanda} \citep{duran08jcim} and 76-dimensional {\it VolSurf} \citep{cruciani00ejps}. In order to calculate the similarity between drugs, we use a Gaussian kernel whose width parameter is selected as the square root of the dimensionality of the data points.

We repeat the same experimental procedure as in the previous experiment with one minor change only. We perform simulations with 16 different numbers of components, i.e., $R \in \{5, 10, \dots, 80\}$, due to the larger size of the interaction network.

We compare four different ways of including the drug property data. \texttt{Amanda} and \texttt{VolSurf} correspond to using a single view when calculating the kernel between drugs. \texttt{Early} corresponds to concatenating the two views, which is known as {\it early combination} \citep{scholkopf04book}, before calculating the kernel between drugs. \texttt{MKL} corresponds to calculating two different kernels between drugs and combining them with our kernel combination approach.

\begin{figure}[!htb]
	\centering
	\includegraphics[scale=0.40]{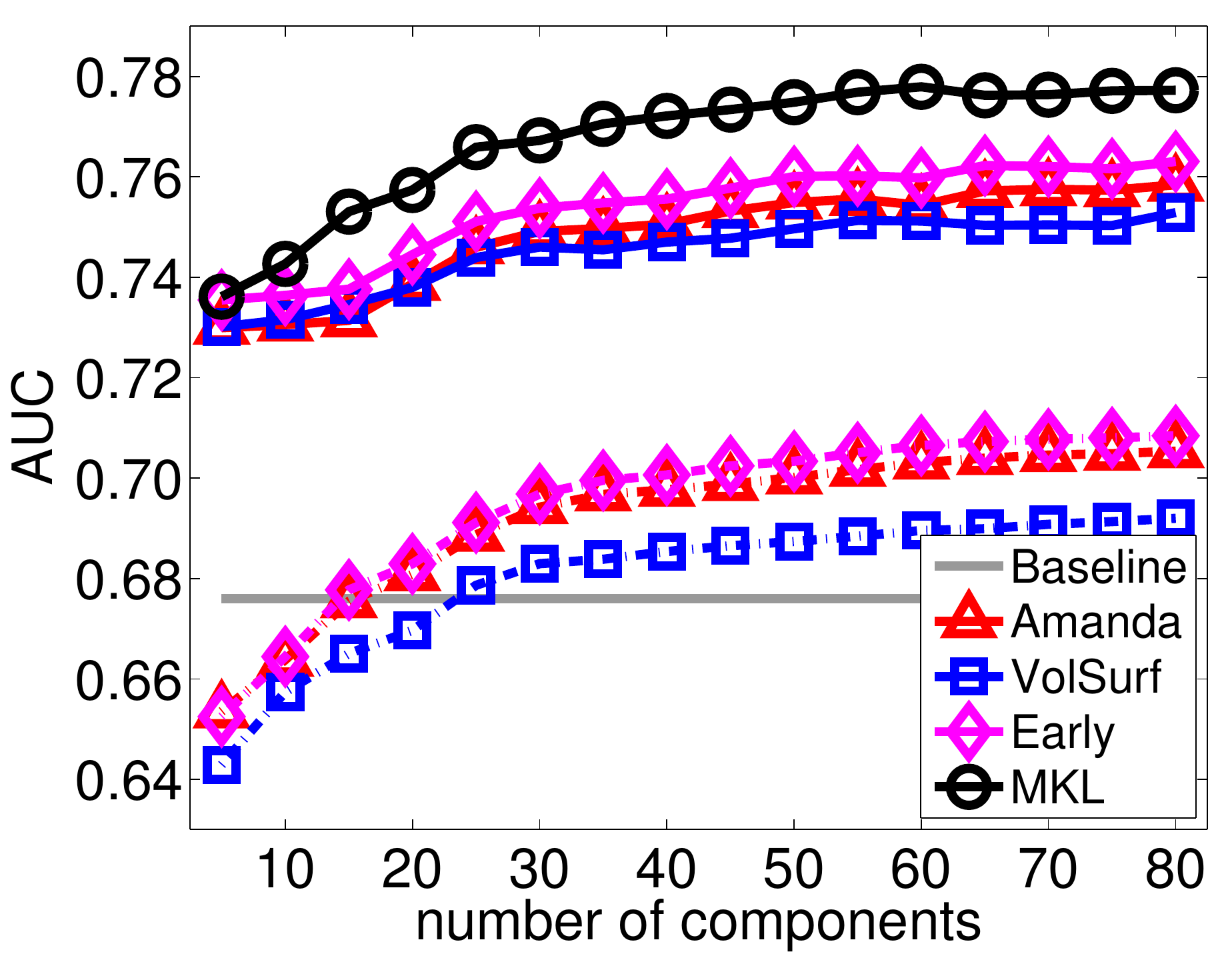}
	\vskip -0.1in
	\caption{Average prediction performances (area under ROC curve, AUC) on the drug--protein data set of \citet{khan12bmcbio}. Gray solid line: \texttt{Baseline}; solid lines: \texttt{KBMF} with binary outputs; dash-dotted lines \texttt{KPMF}.} \label{fig:drug_results}
\end{figure}

The overall ordering of the results of the different matrix factorization methods is the same as in the previous case study (Figure~\ref{fig:drug_results}). The results of \texttt{KBMF} with real-valued outputs, which are omitted not to clutter the figure, are in between \texttt{KPMF} and \texttt{KBMF} with binary outputs. The \texttt{KPMF} outperforms \texttt{Baseline} after 20 components, whereas \texttt{KBMF} is consistently better (by at least four percentage units) than \texttt{KPMF} for all single-kernel scenarios and the differences are statistically significant (paired $t$-test, $p < 0.01$). \texttt{KBMF} with five components is already better than \texttt{Baseline} for all scenarios.

For \texttt{KBMF} with binary outputs, we see that \texttt{Amanda} and \texttt{VolSurf} are significantly better than \texttt{Baseline} and obtain similar prediction performances. \texttt{Early} outperforms \texttt{Amanda} and \texttt{VolSurf} with a slight margin, whereas \texttt{MKL} obtains consistently better results than all the other scenarios after five components.

\begin{figure}[!htb]
	\centering
	\includegraphics[scale=0.40]{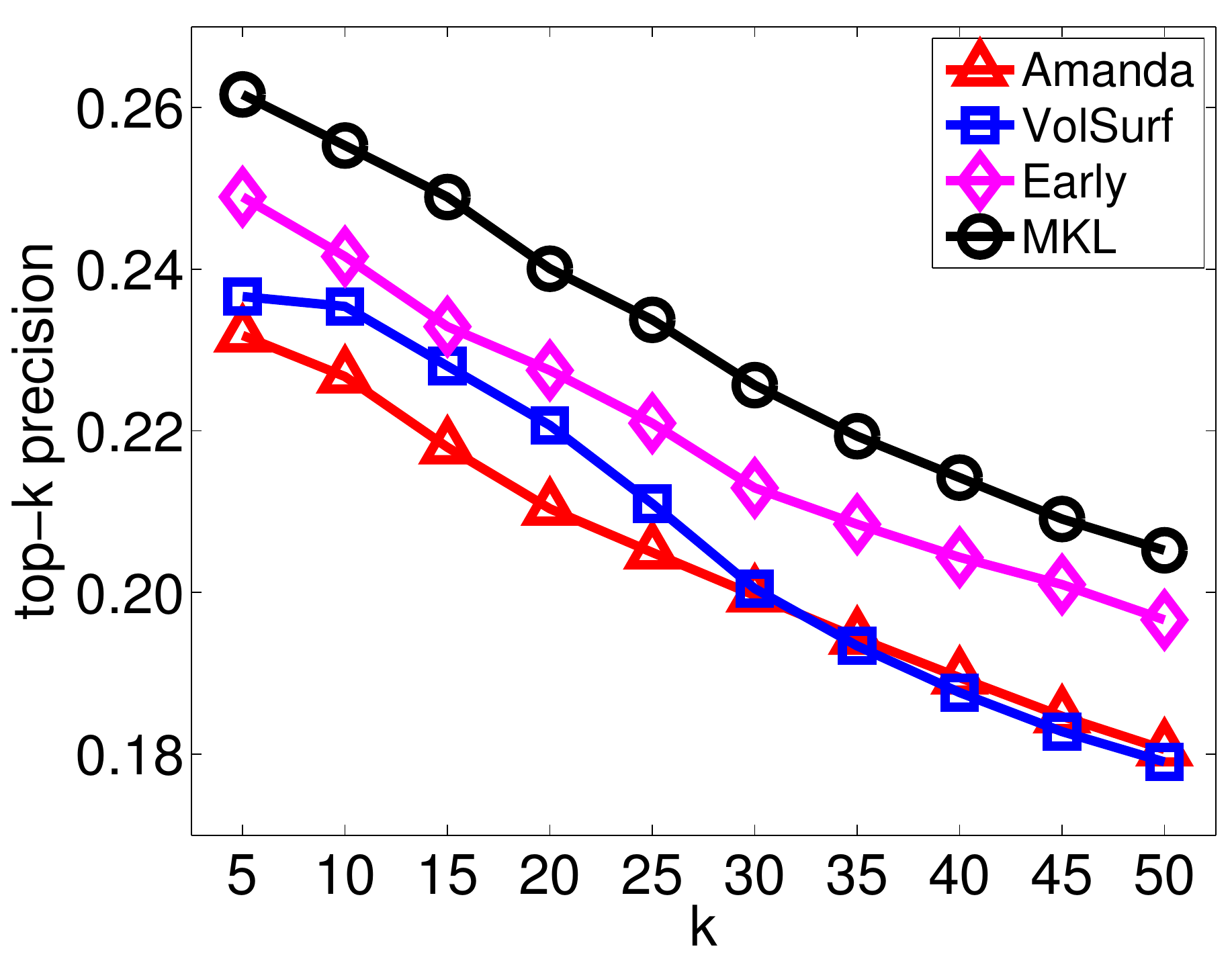}
	\vskip -0.1in
	\caption{Average precision (over query drugs) of retrieval as a function of number $k$ of retrieved drugs. See the text for details.} \label{fig:retrieval_results}
\end{figure}

Our method can also be interpreted as a metric learning algorithm since each kernel function can be converted into a distance metric. We test this property on the task of finding or {\it retrieving} drugs with similar functions. The idea is that since the targets are centrally important for the action mechanisms of drugs, a metric useful for predicting targets could be useful for retrieval of drugs as well. As the ground truth for relevance we use a standard therapeutic classification of the drugs according to the organ or system on which they act and/or their chemical characteristics (not used during learning); drugs having the same class are considered relevant. Figure~\ref{fig:retrieval_results} gives the precision at top-$k$ retrieved drugs, when each drug in turn is used as the query and the rest of the 855 drugs are retrieved in the order of similarity according to the learned metric. \texttt{Early} is better than \texttt{Amanda} and \texttt{VolSurf}, and \texttt{MKL} is the best. This shows that our method is able to learn a kernel function between drugs that is better for retrieval than the kernels either on single or concatenated views.

\begin{table*}[!t]
	\caption{Classification performances (i.e., Hamming loss values) on the multilabel classification data sets. Here, $N_{\texttt{train}}$, $N_{\texttt{test}}$, $D$, and $L$ denote the numbers of training instances, test instances, features, and labels, respectively. The first three data sets are obtained from \burl{http://mulan.sourceforge.net/datasets.html}, whereas the remaining 11 data sets are obtained from \burl{http://www.kecl.ntt.co.jp/as/members/ueda/yahoo.tar.gz}. The figures for comparison algorithms are taken from \citet{zhang12ijcai} and the best result for each data set is marked in bold face.} \label{tab:multilabel}
	\begin{center}
		\small			
		\begin{tabular}{|l|cccc|cccccc|}
			\hline
			\texttt{Data Set} & $N_{\texttt{train}}$ & $N_{\texttt{test}}$ & $D$ & $L$ & \texttt{KBMF} & \texttt{Zhang's} & \texttt{ML-KNN} & \texttt{RML} & \texttt{Tang's} & \texttt{RankSVM} \\
			\hline
			\texttt{Emotions}      & \phantom{0}391 & \phantom{0}202 & \phantom{00}72 & \phantom{0}6 & \bf 0.176 & 0.195 & 0.202 & 0.241 & 0.240 & 0.234 \\
			\texttt{Scene}         & 1211 & 1196 & \phantom{0}294 & \phantom{0}6 & \bf 0.086 & 0.089 & 0.099 & 0.109 & 0.130 & 0.127 \\
			\texttt{Yeast}         & 1500 & \phantom{0}917 & \phantom{0}103 & 14 & \bf 0.189 & 0.196 & 0.195 & 0.204 & 0.190 & 0.201 \\
			\hline
			\texttt{Arts}          & 2000 & 3000 & \phantom{0}462 & 26 & \bf 0.057 & \bf 0.057 & 0.061 & 0.058 & 0.094 & 0.063 \\
			\texttt{Business}      & 2000 & 3000 & \phantom{0}681 & 33 & \bf 0.025 & 0.026 & 0.027 & 0.032 & 0.092 & 0.027 \\
			\texttt{Computers}     & 2000 & 3000 & \phantom{0}640 & 21 & \bf 0.036 & \bf 0.036 & 0.041 & 0.037 & 0.097 & 0.042 \\
			\texttt{Education}     & 2000 & 3000 & \phantom{0}606 & 22 & 0.039 & \bf 0.038 & 0.039 & 0.050 & \bf 0.038 & 0.048 \\
			\texttt{Entertainment} & 2000 & 3000 & \phantom{0}743 & 40 & \bf 0.046 & 0.055 & 0.063 & 0.059 & 0.053 & 0.062 \\
			\texttt{Health}        & 2000 & 3000 & \phantom{0}636 & 27 & \bf 0.036 & 0.037 & 0.047 & 0.041 & 0.222 & 0.042 \\
			\texttt{Recreation}    & 2000 & 3000 & \phantom{0}438 & 30 & \bf 0.044 & 0.057 & 0.062 & 0.057 & 0.057 & 0.064 \\
			\texttt{Reference}     & 2000 & 3000 & \phantom{0}550 & 33 & 0.027 & \bf 0.025 & 0.032 & 0.027 & 0.087 & 0.034 \\
			\texttt{Science}       & 2000 & 3000 & \phantom{0}612 & 32 & 0.032 & \bf 0.031 & 0.033 & 0.051 & 0.057 & 0.038 \\
			\texttt{Social}        & 2000 & 3000 & \phantom{0}793 & 33 & 0.022 & \bf 0.021 & 0.022 & 0.101 & 0.072 & 0.027 \\
			\texttt{Society}       & 2000 & 3000 & 1047 & 39 & \bf 0.038 & 0.052 & 0.054 & 0.096 & 0.056 & 0.060 \\
			\hline
                 \multicolumn{5}{|r|}{\texttt{Average Rank}} & \bf 1.536 & 1.964 & 3.750 & 4.464 & 4.607 & 4.679 \\
			\hline
		\end{tabular}
	\end{center}
	\vskip -0.2in
\end{table*}

\subsection{Multilabel Classification Data Sets}
In multilabel learning, each sample is associated with a set of labels instead of just a single label. Multilabel classification can be cast into our formulation as follows: Samples and labels are assumed to be from domains $\mathcal{X}$ and $\mathcal{Z}$, respectively. Class membership matrix corresponds to target label matrix $\mat{Y}$ in our model. Our method allows us to integrate side information about samples and labels in the form of kernel matrices. For example, we can exploit the correlation between labels by integrating a kernel calculated over them into the model.

We compare our algorithm \texttt{KBMF} with five state-of-the-art multilabel learning algorithms, namely, (i)~\texttt{RankSVM} \citep{elisseeff02nips}, (ii)~\texttt{ML-KNN} \citep{zhang07pr}, (iii) \texttt{Tang's} \citep{tang09ijcai}, (iv)~\texttt{RML} \citep{petterson10nips}, and (v)~\texttt{Zhang's} \citep{zhang12ijcai}. We perform experiments on 14 benchmark multilabel classification data sets whose characteristics are given in Table~\ref{tab:multilabel}.

For \texttt{KBMF}, the similarities between samples are measured with five different Gaussian kernels whose widths are selected as $\sqrt{D/4}$, $\sqrt{D/2}$, $\sqrt{D}$, $\sqrt{2D}$, and $\sqrt{4D}$, whereas the similarity between labels is measured with the Jaccard index over the labels of training samples. The number of components $R$ is selected from $\{1,\dots,\min(L, 15)\}$ according to training performance.

Table~\ref{tab:multilabel} reports the classification results on multilabel data sets. \texttt{KBMF} obtains the best results on 10 out of 14 data sets, whereas it obtains the second best results on the remaining four data sets. These results validate the suitability of our framework to multilabel learning.

%%%%%%%%%%%%%%%%%%%%%%%%%%%%%%%%%%%%%%%%%%%%%%%%%%%%%%%%%%%%%%%%%%%%%%%%%%%%%%%%%%%%%%%%%%%%%%%%%%%%

\section{Discussion} \label{sec:discussion}
We introduce a kernelized Bayesian matrix factorization method that can make use of multiple side information sources about the objects (both rows and columns) and be applied in various scenarios including recommender systems, interaction network modeling, and multilabel learning. Our two main contributions are: (i)~formulating an efficient variational approximation scheme for inference with the help of a novel fully conjugate probabilistic model and (ii)~coupling matrix factorization with multiple kernel learning to integrate multiple side information sources into the model. In contrast to the earlier kernelized probabilistic matrix factorization method of \citet{zhou12sdm}, for our probabilistic model, it is possible to derive a computationally feasible fully Bayesian treatment. We illustrate the usefulness of the method on one toy data set, two molecular biological data sets, and 14 multilabel classification data sets.

An interesting topic for future research is to optimize the dimensionality of the latent components using a Bayesian model selection procedure. For example, we can share the same set of precision priors for the projection matrices and determine the dimensionality using {\it automatic relevance determination} \citep{neal96book}.

%%%%%%%%%%%%%%%%%%%%%%%%%%%%%%%%%%%%%%%%%%%%%%%%%%%%%%%%%%%%%%%%%%%%%%%%%%%%%%%%%%%%%%%%%%%%%%%%%%%%

{\bf Acknowledgments.} This work was financially supported by the Academy of Finland (Finnish Centre of Excellence in Computational Inference Research COIN, grant no 251170; Computational Modeling of the Biological Effects of Chemicals, grant no 140057) and the Finnish Graduate School in Computational Sciences (FICS).

%%%%%%%%%%%%%%%%%%%%%%%%%%%%%%%%%%%%%%%%%%%%%%%%%%%%%%%%%%%%%%%%%%%%%%%%%%%%%%%%%%%%%%%%%%%%%%%%%%%%

\renewcommand\refname{\normalsize References}
{\small
\bibliography{kbmf2mkl}
\bibliographystyle{icml2013}}
%\end{document}

%%%%%%%%%%%%%%%%%%%%%%%%%%%%%%%%%%%%%%%%%%%%%%%%%%%%%%%%%%%%%%%%%%%%%%%%%%%%%%%%%%%%%%%%%%%%%%%%%%%%

\onecolumn
\appendix
\allowdisplaybreaks[4]

\section*{\Large \centering Supplementary Material}
In this supplementary material, we provide details about our method and its two variants: (\ref{app:kbmf2mkl_updates}) variational updates for the model presented in the main paper, (\ref{app:kbmf2k}) variant for using a single kernel function on each domain instead of the multiple kernels, and (\ref{app:kbmf2mkl_regression}) variant for predicting real-valued outputs instead of the binary outputs.

%%%%%%%%%%%%%%%%%%%%%%%%%%%%%%%%%%%%%%%%%%%%%%%%%%%%%%%%%%%%%%%%%%%%%%%%%%%%%%%%%%%%%%%%%%%%%%%%%%%%

\section{Variational Updates for Kernelized Bayesian Matrix Factorization with Twin Multiple Kernel Learning} \label{app:kbmf2mkl_updates}
The approximate posterior distributions of the dimensionality reduction part can be found as
\begin{align*}
	q(\mat{\Lambda}_{\texttt{x}}) &= \prod \limits_{i = 1}^{N_{\texttt{x}}} \prod \limits_{s = 1}^{R} \mathcal{G}\left(\lambda_{\texttt{x},s}^{i}; \alpha_{\lambda} + \dfrac{1}{2}, \left(\dfrac{1}{\beta_{\lambda}} + \dfrac{\widetilde{(a_{\texttt{x},s}^{i})^{2}}}{2} \right)^{-1}\right) \\
	q(\mat{A}_{\texttt{x}}) &= \prod \limits_{s = 1}^{R} \mathcal{N}\left(\vec{a}_{\texttt{x},s}; \Sigma(\vec{a}_{\texttt{x},s}) \sum \limits_{m = 1}^{P_{\texttt{x}}} \dfrac{\mat{K}_{\texttt{x},m} \widetilde{(\vec{g}_{\texttt{x},m}^{s})^{\top}}}{\sigma_{g}^{2}}, \left(\diag(\widetilde{\vec{\lambda}_{\texttt{x}}^{s}}) + \sum \limits_{m = 1}^{P_{\texttt{x}}} \dfrac{\mat{K}_{\texttt{x},m} \mat{K}_{\texttt{x},m}^{\top}}{\sigma_{g}^{2}}\right)^{-1}\right)
\end{align*}
where the tilde notation denotes the posterior expectations as usual, i.e., $\widetilde{f(\vec{\tau})} = \E_{ q(\vec{\tau})}[f(\vec{\tau})]$.

The kernel-specific components have the following approximate posterior distribution:
\begin{align*}
	q(\{\mat{G}_{\texttt{x},m}\}_{m = 1}^{P_{\texttt{x}}}) = \prod \limits_{m = 1}^{P_{\texttt{x}}} \prod \limits_{i = 1}^{N_{\texttt{x}}} \mathcal{N}\left(\vec{g}_{\texttt{x},m,i}; \Sigma(\vec{g}_{\texttt{x},m,i}) \left(\dfrac{\widetilde{\mat{A}_{\texttt{x}}^{\top}} \vec{k}_{\texttt{x},m,i}}{\sigma_{g}^{2}} + \dfrac{\widetilde{e_{\texttt{x},m}} \widetilde{\vec{h}_{\texttt{x},i}}}{\sigma_{h}^{2}} - \sum \limits_{o \neq m} \dfrac{\widetilde{e_{\texttt{x},m}e_{\texttt{x},o}} \widetilde{\vec{g}_{\texttt{x},o,i}}}{\sigma_{h}^{2}} \right), \left(\dfrac{\mat{I}}{\sigma_{g}^{2}} + \dfrac{\widetilde{e_{\texttt{x}, m}^{2}} \mat{I}}{\sigma_{h}^{2}}\right)^{-1}\right)
\end{align*}
where the mean and covariance parameters are affected by the kernel weights, the composite components, and other kernel-specific components in addition to the projection matrix and the corresponding kernel matrix. 

The approximate posterior distributions of the multiple kernel learning part can be found as
\begin{align*}
	q(\vec{\eta}_{\texttt{x}}) &= \prod \limits_{m = 1}^{P_{\texttt{x}}} \mathcal{G}\left(\eta_{\texttt{x},m}; \alpha_{\eta} + \dfrac{1}{2}, \left(\dfrac{1}{\beta_{\eta}} + \dfrac{\widetilde{e_{\texttt{x},m}^{2}}}{2} \right)^{-1}\right) \\
	q(\vec{e}_{\texttt{x}}) &= \mathcal{N}\left(\vec{e}_{\texttt{x}}; \Sigma(\vec{e}_{\texttt{x}}) \begin{bmatrix}\dfrac{\widetilde{\mat{G}_{\texttt{x},m}^{\top}} \widetilde{\mat{H}_{\texttt{x}}}}{\sigma_{h}^{2}} \end{bmatrix}_{m = 1}^{P_{\texttt{x}}}, \left(\diag(\widetilde{\vec{\eta}_{\texttt{x}}}) + \begin{bmatrix}\dfrac{\widetilde{\mat{G}_{\texttt{x},m}^{\top}\mat{G}_{\texttt{x},o}}}{\sigma_{h}^{2}} \end{bmatrix}_{m = 1, o = 1}^{P_{\texttt{x}}, P_{\texttt{x}}}\right)^{-1} \right)
\end{align*}
where the mean and covariance parameters of the kernel weights are calculated using the kernel-specific and composite components.

The composite components have the following approximate posterior distribution: 
\begin{align*}
	q(\mat{H}_{\texttt{x}}) = \prod \limits_{i = 1}^{N_{\texttt{x}}} \mathcal{N}\left(\vec{h}_{\texttt{x},i}; \Sigma(\vec{h}_{\texttt{x},i}) \left(\sum \limits_{m = 1}^{P_{\texttt{x}}} \dfrac{\widetilde{e_{\texttt{x},m}} \widetilde{\vec{g}_{\texttt{x},m,i}}}{\sigma_{h}^{2}} + \widetilde{\mat{H}_{\texttt{z}}} \widetilde{(\vec{f}^{i})^{\top}} \right), \left(\dfrac{\mat{I}}{\sigma_{h}^{2}} + \widetilde{\mat{H}_{\texttt{z}} \mat{H}_{\texttt{z}}^{\top}} \right)^{-1}\right)
\end{align*}
where it can be seen that the inference transfers information between the two domains. Note that the composite components of each domain are the only random variables that have an effect on the other domain, i.e., only the $\mat{H}_{\texttt{z}}$ variables of domain $\mathcal{Z}$ are used when updating the random variables of the domain $\mathcal{X}$.

The approximate posterior distribution of the predicted outputs is a product of truncated normals:
\begin{align*}
	q(\mat{F}) &= \prod \limits_{i = 1}^{N_{\texttt{x}}} \prod \limits_{j = 1}^{N_{\texttt{z}}} \mathcal{TN}(f_{j}^{i}; \widetilde{\vec{h}_{\texttt{x},i}^{\top}} \widetilde{\vec{h}_{\texttt{z},j}}, 1, f_{j}^{i} y_{j}^{i} > \nu).
\end{align*}
We need to find the posterior expectation of $\mat{F}$ to update the approximate posterior distributions of the composite components. Fortunately, the truncated normal distribution has a closed-form formula for its expectation.

%%%%%%%%%%%%%%%%%%%%%%%%%%%%%%%%%%%%%%%%%%%%%%%%%%%%%%%%%%%%%%%%%%%%%%%%%%%%%%%%%%%%%%%%%%%%%%%%%%%%

\section{Kernelized Bayesian Matrix Factorization with Twin Kernels} \label{app:kbmf2k}
\begin{figure}[!ht]
	\centering
	\includegraphics[scale=0.8]{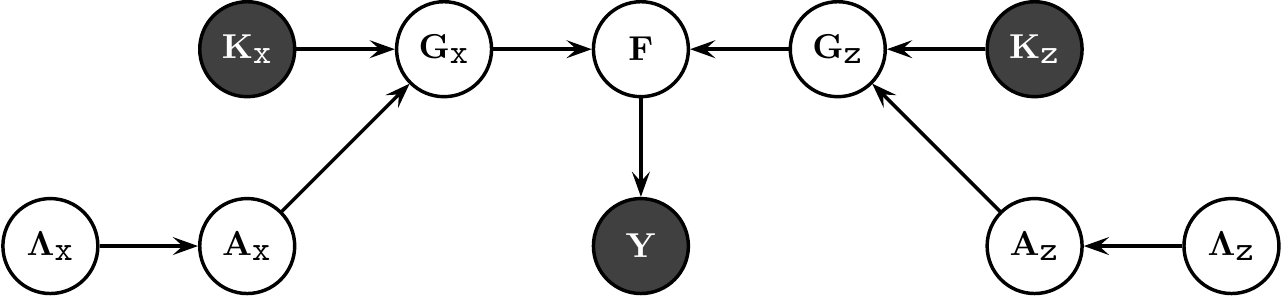}
	\vskip -0.1in
	\caption{Graphical model of kernelized Bayesian matrix factorization with twin kernels.} \label{fig:kbmf2k_graphical}
\end{figure}

We formulate a simplified probabilistic model, called {\it kernelized Bayesian matrix factorization with twin kernels} (KBMF2K), for the case with a single kernel function for each domain. Figure~\ref{fig:kbmf2k_graphical} shows the graphical model of KBMF2K with latent variables and their corresponding priors.

The distributional assumptions of the simplified model are
\begin{alignat*}{4}
	\lambda_{\texttt{x},s}^{i} &\sim \mathcal{G}(\lambda_{\texttt{x},s}^{i}; \alpha_{\lambda}, \beta_{\lambda}) &&\largespace \forall (i, s) \\
	a_{\texttt{x},s}^{i} | \lambda_{\texttt{x},s}^{i} &\sim \mathcal{N}(a_{\texttt{x},s}^{i}; 0, (\lambda_{\texttt{x},s}^{i})^{-1}) &&\largespace \forall (i, s) \\
	g_{\texttt{x},i}^{s} | \vec{a}_{\texttt{x},s}, \vec{k}_{\texttt{x},i} &\sim \mathcal{N}(g_{\texttt{x},i}^{s}; \vec{a}_{\texttt{x},s}^{\top} \vec{k}_{\texttt{x},i}, \sigma_{g}^{2}) &&\largespace \forall (s, i) \\
	f_{j}^{i} | \vec{g}_{\texttt{x},i}, \vec{g}_{\texttt{z},j} &\sim \mathcal{N}(f_{j}^{i}; \vec{g}_{\texttt{x},i}^{\top} \vec{g}_{\texttt{z},j}, 1) &&\largespace \forall (i, j) \\
	y_{j}^{i} | f_{j}^{i} &\sim \delta(y_{j}^{i}; f_{j}^{i} y_{j}^{i} > \nu) &&\largespace \forall (i, j).
\end{alignat*}

As short-hand notations, all hyper-parameters in the model are denoted by $\vec{\zeta} = \{\alpha_{\lambda}, \beta_{\lambda}, \sigma_{g}, \nu\}$, all prior variables by $\mat{\Xi} = \{\mat{\Lambda}_{\texttt{x}}, \mat{\Lambda}_{\texttt{z}}\}$, and the remaining random variables by $\mat{\Theta} = \{\mat{A}_{\texttt{x}}, \mat{A}_{\texttt{z}}, \mat{F}, \mat{G}_{\texttt{x}}, \mat{G}_{\texttt{z}}\}$. We again omit the dependence on $\vec{\zeta}$ for clarity. We can write the factorized variational approximation as
\begin{align*}
	p(\mat{\Theta}, \mat{\Xi} | \mat{K}_{\texttt{x}}, \mat{K}_{\texttt{z}}, \mat{Y}) \approx q(\mat{\Theta}, \mat{\Xi}) = q(\mat{\Lambda}_{\texttt{x}}) q(\mat{A}_{\texttt{x}}) q(\mat{G}_{\texttt{x}}) q(\mat{\Lambda}_{\texttt{z}}) q(\mat{A}_{\texttt{z}}) q(\mat{G}_{\texttt{z}}) q(\mat{F})
\end{align*}
and define each factor in the ensemble just like its full conditional:
\begin{align*}
	q(\mat{\Lambda}_{\texttt{x}}) &= \prod \limits_{i = 1}^{N_{\texttt{x}}} \prod \limits_{s = 1}^{R} \mathcal{G}(\lambda_{\texttt{x},s}^{i}; \alpha(\lambda_{\texttt{x},s}^{i}), \beta(\lambda_{\texttt{x},s}^{i})) \\
	q(\mat{A}_{\texttt{x}}) &= \prod \limits_{s = 1}^{R} \mathcal{N}(\vec{a}_{\texttt{x},s}; \mu(\vec{a}_{\texttt{x},s}), \Sigma(\vec{a}_{\texttt{x},s})) \\
	q(\mat{G}_{\texttt{x}}) &= \prod \limits_{i = 1}^{N_{\texttt{x}}} \mathcal{N}(\vec{g}_{\texttt{x},i}; \mu(\vec{g}_{\texttt{x},i}), \Sigma(\vec{g}_{\texttt{x},i})) \\
	q(\mat{F}) &= \prod \limits_{i = 1}^{N_{\texttt{x}}} \prod \limits_{j = 1}^{N_{\texttt{z}}} \mathcal{TN}(f_{j}^{i}; \mu(f_{j}^{i}), \Sigma(f_{j}^{i}), \rho(f_{j}^{i})).
\end{align*}

We can bound the marginal likelihood using Jensen's inequality:
\begin{align*}
	\log p(\mat{Y} | \mat{K}_{\texttt{x}}, \mat{K}_{\texttt{z}}) \geq \E_{q(\mat{\Theta}, \mat{\Xi})}[\log p(\mat{Y}, \mat{\Theta}, \mat{\Xi} | \mat{K}_{\texttt{x}}, \mat{K}_{\texttt{z}})] - \E_{q(\mat{\Theta}, \mat{\Xi})}[\log q(\mat{\Theta}, \mat{\Xi})]
\end{align*}
and optimize this bound by maximizing with respect to each factor separately until convergence. The approximate posterior distribution of a specific factor $\vec{\tau}$ can be found as
\begin{align*}
	q(\vec{\tau}) &\propto \exp(\E_{q(\{\mat{\Theta}, \mat{\Xi}\} \backslash \vec{\tau} )}[\log p(\mat{Y}, \mat{\Theta}, \mat{\Xi} | \mat{K}_{\texttt{x}}, \mat{K}_{\texttt{z}})]).
\end{align*}

The approximate posterior distributions of the ensemble can be found as
\begin{align*}
	q(\mat{\Lambda}_{\texttt{x}}) &= \prod \limits_{i = 1}^{N_{\texttt{x}}} \prod \limits_{s = 1}^{R} \mathcal{G}\left(\lambda_{\texttt{x},s}^{i}; \alpha_{\lambda} + \dfrac{1}{2}, \left(\dfrac{1}{\beta_{\lambda}} + \dfrac{\widetilde{(a_{\texttt{x},s}^{i})^{2}}}{2} \right)^{-1}\right) \\
	q(\mat{A}_{\texttt{x}}) &= \prod \limits_{s = 1}^{R} \mathcal{N}\left(\vec{a}_{\texttt{x},s}; \Sigma(\vec{a}_{\texttt{x},s}) \dfrac{\mat{K}_{\texttt{x}} \widetilde{(\vec{g}_{\texttt{x}}^{s})^{\top}}}{\sigma_{g}^{2}}, \left(\diag(\widetilde{\vec{\lambda}_{\texttt{x}}^{s}}) + \dfrac{\mat{K}_{\texttt{x},m} \mat{K}_{\texttt{x},m}^{\top}}{\sigma_{g}^{2}}\right)^{-1}\right) \\
	q(\mat{G}_{\texttt{x}}) &= \prod \limits_{i = 1}^{N_{\texttt{x}}} \mathcal{N}\left(\vec{g}_{\texttt{x},i}; \Sigma(\vec{g}_{\texttt{x},i}) \left(\dfrac{\widetilde{\mat{A}_{\texttt{x}}^{\top}} \vec{k}_{\texttt{x},i}}{\sigma_{g}^{2}} + \widetilde{\mat{G}_{\texttt{z}}} \widetilde{(\vec{f}^{i})^{\top}} \right), \left(\dfrac{\mat{I}}{\sigma_{g}^{2}} + \widetilde{\mat{G}_{\texttt{z}} \mat{G}_{\texttt{z}}^{\top}} \right)^{-1}\right) \\
	q(\mat{F}) &= \prod \limits_{i = 1}^{N_{\texttt{x}}} \prod \limits_{j = 1}^{N_{\texttt{z}}} \mathcal{TN}(f_{j}^{i}; \widetilde{\vec{g}_{\texttt{x},i}^{\top}} \widetilde{\vec{g}_{\texttt{z},j}}, 1, f_{j}^{i} y_{j}^{i} > \nu).
\end{align*}

%%%%%%%%%%%%%%%%%%%%%%%%%%%%%%%%%%%%%%%%%%%%%%%%%%%%%%%%%%%%%%%%%%%%%%%%%%%%%%%%%%%%%%%%%%%%%%%%%%%%

\section{Kernelized Bayesian Matrix Factorization with Twin Multiple Kernel Learning for Real-Valued Outputs} \label{app:kbmf2mkl_regression}
\begin{figure}[!ht]
	\centering
	\includegraphics[scale=0.8]{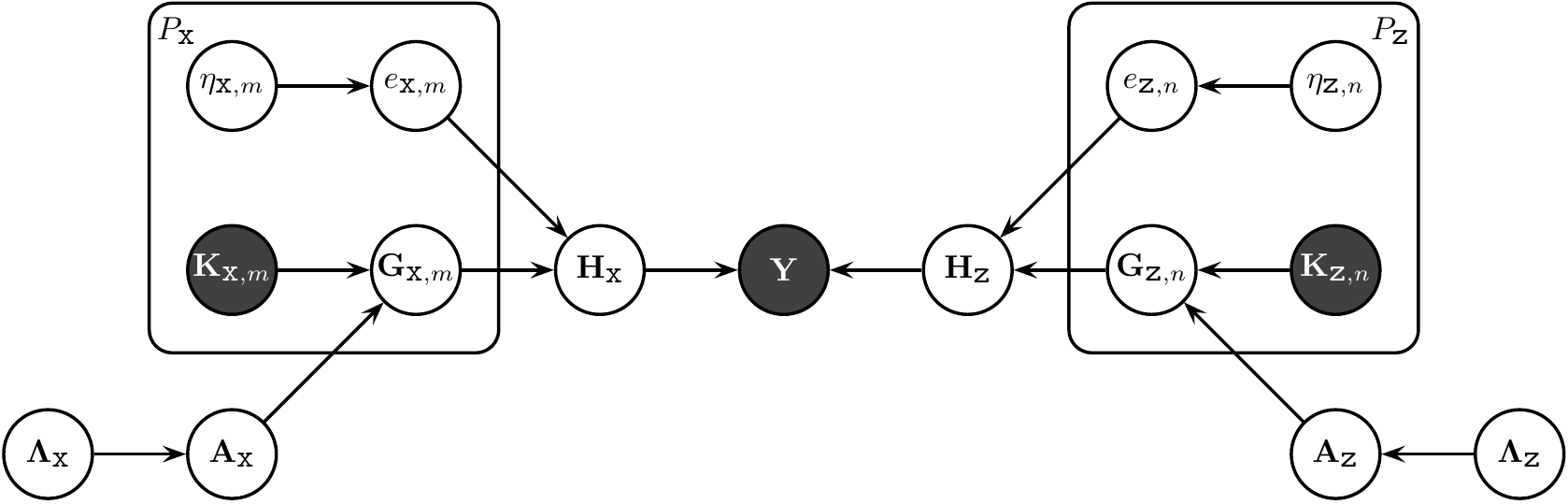}
	\vskip -0.1in
	\caption{Graphical model of kernelized Bayesian matrix factorization with twin multiple kernel learning for real-valued outputs.} \label{fig:kbmf2mkl_regression_graphical}	
\end{figure}

We modify our proposed model for binary-valued outputs to also handle real-valued outputs. Figure~\ref{fig:kbmf2mkl_regression_graphical} illustrates the graphical model of the modified {\it kernelized Bayesian matrix factorization with twin multiple kernel learning} (KBMF2MKL) with latent variables and their corresponding priors.

The distributional assumptions of the modified KBMF2MKL model are
\begin{alignat*}{4}
	\lambda_{\texttt{x},s}^{i} &\sim \mathcal{G}(\lambda_{\texttt{x},s}^{i}; \alpha_{\lambda}, \beta_{\lambda}) &&\largespace \forall (i, s) \\
	a_{\texttt{x},s}^{i} | \lambda_{\texttt{x},s}^{i} &\sim \mathcal{N}(a_{\texttt{x},s}^{i}; 0, (\lambda_{\texttt{x},s}^{i})^{-1}) &&\largespace \forall (i, s) \\
	g_{\texttt{x},m,i}^{s} | \vec{a}_{\texttt{x},s}, \vec{k}_{\texttt{x},m,i} &\sim \mathcal{N}(g_{\texttt{x},m,i}^{s}; \vec{a}_{\texttt{x},s}^{\top} \vec{k}_{\texttt{x},m,i}, \sigma_{g}^{2}) &&\largespace \forall (m, s, i) \\
	\eta_{\texttt{x},m} &\sim \mathcal{G}(\eta_{\texttt{x},m}; \alpha_{\eta}, \beta_{\eta}) &&\largespace \forall m \\\
	e_{\texttt{x},m} | \eta_{\texttt{x},m} &\sim \mathcal{N}(e_{\texttt{x},m}; 0, \eta_{\texttt{x},m}^{-1}) &&\largespace \forall m \\
	h_{\texttt{x},i}^{s} | \{e_{\texttt{x}, m}, g_{\texttt{x},m,i}^{s}\}_{m = 1}^{P_{\texttt{x}}} &\sim \mathcal{N}\left(h_{\texttt{x},i}^{s}; \sum \limits_{m = 1}^{P_{\texttt{x}}} e_{\texttt{x},m} g_{\texttt{x},m,i}^{s}, \sigma_{h}^{2}\right) &&\largespace \forall (s, i) \\
	y_{j}^{i} | \vec{h}_{\texttt{x},i}, \vec{h}_{\texttt{z},j} &\sim \mathcal{N}(y_{j}^{i}; \vec{h}_{\texttt{x},i}^{\top} \vec{h}_{\texttt{z},j}, \sigma_{y}^{2}) &&\largespace \forall (i, j).
\end{alignat*}

As short-hand notations, all hyper-parameters in the model are denoted by $\vec{\zeta} = \{\alpha_{\eta}, \beta_{\eta}, \alpha_{\lambda}, \beta_{\lambda}, \sigma_{g}, \sigma_{h}, \sigma_{y}\}$, all prior variables by $\mat{\Xi} = \{\vec{\eta}_{\texttt{x}}, \vec{\eta}_{\texttt{z}}, \mat{\Lambda}_{\texttt{x}}, \mat{\Lambda}_{\texttt{z}}\}$, and the remaining random variables by $\mat{\Theta} = \{\mat{A}_{\texttt{x}}, \mat{A}_{\texttt{z}}, \vec{e}_{\texttt{x}}, \vec{e}_{\texttt{z}}, \{\mat{G}_{\texttt{x},m}\}_{m = 1}^{P_{\texttt{x}}}, \{\mat{G}_{\texttt{z},n}\}_{n = 1}^{P_{\texttt{z}}}, \mat{H}_{\texttt{x}}, \mat{H}_{\texttt{z}}\}$. We again omit the dependence on $\vec{\zeta}$ for clarity. We can write the factorized variational approximation as
\begin{multline*}
	p(\mat{\Theta}, \mat{\Xi} | \{\mat{K}_{\texttt{x},m}\}_{m = 1}^{P_{\texttt{x}}}, \{\mat{K}_{\texttt{z},n}\}_{n = 1}^{P_{\texttt{z}}}, \mat{Y}) \approx q(\mat{\Theta}, \mat{\Xi}) = \\ q(\mat{\Lambda}_{\texttt{x}}) q(\mat{A}_{\texttt{x}}) q(\{\mat{G}_{\texttt{x},m}\}_{m = 1}^{P_{\texttt{x}}}) q(\vec{\eta}_{\texttt{x}}) q(\vec{e}_{\texttt{x}}) q(\mat{H}_{\texttt{x}}) q(\mat{\Lambda}_{\texttt{z}}) q(\mat{A}_{\texttt{z}}) q(\{\mat{G}_{\texttt{z},n}\}_{n = 1}^{P_{\texttt{z}}}) q(\vec{\eta}_{\texttt{z}}) q(\vec{e}_{\texttt{z}}) q(\mat{H}_{\texttt{z}})
\end{multline*}
and define each factor in the ensemble just like its full conditional:
\begin{align*}
	q(\mat{\Lambda}_{\texttt{x}}) &= \prod \limits_{i = 1}^{N_{\texttt{x}}} \prod \limits_{s = 1}^{R} \mathcal{G}(\lambda_{\texttt{x},s}^{i}; \alpha(\lambda_{\texttt{x},s}^{i}), \beta(\lambda_{\texttt{x},s}^{i})) \\
	q(\mat{A}_{\texttt{x}}) &= \prod \limits_{s = 1}^{R} \mathcal{N}(\vec{a}_{\texttt{x},s}; \mu(\vec{a}_{\texttt{x},s}), \Sigma(\vec{a}_{\texttt{x},s})) \\
	q(\{\mat{G}_{\texttt{x},m}\}_{m = 1}^{P_{\texttt{x}}}) &= \prod \limits_{m = 1}^{P_{\texttt{x}}} \prod \limits_{i = 1}^{N_{\texttt{x}}} \mathcal{N}(\vec{g}_{\texttt{x},m,i}; \mu(\vec{g}_{\texttt{x},m,i}), \Sigma(\vec{g}_{\texttt{x},m,i})) \\
	q(\vec{e}_{\texttt{x}}) &= \mathcal{N}(\vec{e}_{\texttt{x}}; \mu(\vec{e}_{\texttt{x}}), \Sigma(\vec{e}_{\texttt{x}})) \\
	q(\vec{\eta}_{\texttt{x}}) &= \prod \limits_{m = 1}^{P_{\texttt{x}}} \mathcal{G}(\eta_{\texttt{x},m}; \alpha(\eta_{\texttt{x},m}), \beta(\eta_{\texttt{x},m})) \\
	q(\mat{H}_{\texttt{x}}) &= \prod \limits_{i = 1}^{N_{\texttt{x}}} \mathcal{N}(\vec{h}_{\texttt{x},i}; \mu(\vec{h}_{\texttt{x},i}), \Sigma(\vec{h}_{\texttt{x},i})).
\end{align*}

We can bound the marginal likelihood using Jensen's inequality:
\begin{align*}
	\log p(\mat{Y} | \{\mat{K}_{\texttt{x},m}\}_{m = 1}^{P_{\texttt{x}}}, \{\mat{K}_{\texttt{z},n}\}_{n = 1}^{P_{\texttt{z}}}) \geq \E_{q(\mat{\Theta}, \mat{\Xi})}[\log p(\mat{Y}, \mat{\Theta}, \mat{\Xi} | \{\mat{K}_{\texttt{x},m}\}_{m = 1}^{P_{\texttt{x}}}, \{\mat{K}_{\texttt{z},n}\}_{n = 1}^{P_{\texttt{z}}})] - \E_{q(\mat{\Theta}, \mat{\Xi})}[\log q(\mat{\Theta}, \mat{\Xi})]
\end{align*}
and optimize this bound by maximizing with respect to each factor separately until convergence. The approximate posterior distribution of a specific factor $\vec{\tau}$ can be found as
\begin{align*}
	q(\vec{\tau}) &\propto \exp(\E_{q(\{\mat{\Theta}, \mat{\Xi}\} \backslash \vec{\tau} )}[\log p(\mat{Y}, \mat{\Theta}, \mat{\Xi} | \{\mat{K}_{\texttt{x},m}\}_{m = 1}^{P_{\texttt{x}}}, \{\mat{K}_{\texttt{z},n}\}_{n = 1}^{P_{\texttt{z}}})]).
\end{align*}

The approximate posterior distributions of the ensemble can be found as
\begin{align*}
	q(\mat{\Lambda}_{\texttt{x}}) &= \prod \limits_{i = 1}^{N_{\texttt{x}}} \prod \limits_{s = 1}^{R} \mathcal{G}\left(\lambda_{\texttt{x},s}^{i}; \alpha_{\lambda} + \dfrac{1}{2}, \left(\dfrac{1}{\beta_{\lambda}} + \dfrac{\widetilde{(a_{\texttt{x},s}^{i})^{2}}}{2} \right)^{-1}\right) \nonumber \\
	q(\mat{A}_{\texttt{x}}) &= \prod \limits_{s = 1}^{R} \mathcal{N}\left(\vec{a}_{\texttt{x},s}; \Sigma(\vec{a}_{\texttt{x},s}) \sum \limits_{m = 1}^{P_{\texttt{x}}} \dfrac{\mat{K}_{\texttt{x},m} \widetilde{(\vec{g}_{\texttt{x},m}^{s})^{\top}}}{\sigma_{g}^{2}}, \left(\diag(\widetilde{\vec{\lambda}_{\texttt{x}}^{s}}) + \sum \limits_{m = 1}^{P_{\texttt{x}}} \dfrac{\mat{K}_{\texttt{x},m} \mat{K}_{\texttt{x},m}^{\top}}{\sigma_{g}^{2}}\right)^{-1}\right) \\
	q(\{\mat{G}_{\texttt{x},m}\}_{m = 1}^{P_{\texttt{x}}}) &= \prod \limits_{m = 1}^{P_{\texttt{x}}} \prod \limits_{i = 1}^{N_{\texttt{x}}} \mathcal{N}\left(\vec{g}_{\texttt{x},m,i}; \Sigma(\vec{g}_{\texttt{x},m,i}) \left(\dfrac{\widetilde{\mat{A}_{\texttt{x}}^{\top}} \vec{k}_{\texttt{x},m,i}}{\sigma_{g}^{2}} + \dfrac{\widetilde{e_{\texttt{x},m}} \widetilde{\vec{h}_{\texttt{x},i}}}{\sigma_{h}^{2}} - \sum \limits_{o \neq m} \dfrac{\widetilde{e_{\texttt{x},m}e_{\texttt{x},o}} \widetilde{\vec{g}_{\texttt{x},o,i}}}{\sigma_{h}^{2}} \right), \left(\dfrac{\mat{I}}{\sigma_{g}^{2}} + \dfrac{\widetilde{e_{\texttt{x}, m}^{2}} \mat{I}}{\sigma_{h}^{2}}\right)^{-1}\right)\\
	q(\vec{\eta}_{\texttt{x}}) &= \prod \limits_{m = 1}^{P_{\texttt{x}}} \mathcal{G}\left(\eta_{\texttt{x},m}; \alpha_{\eta} + \dfrac{1}{2}, \left(\dfrac{1}{\beta_{\eta}} + \dfrac{\widetilde{e_{\texttt{x},m}^{2}}}{2} \right)^{-1}\right) \nonumber \\
	q(\vec{e}_{\texttt{x}}) &= \mathcal{N}\left(\vec{e}_{\texttt{x}}; \Sigma(\vec{e}_{\texttt{x}}) \begin{bmatrix}\dfrac{\widetilde{\mat{G}_{\texttt{x},m}^{\top}} \widetilde{\mat{H}_{\texttt{x}}}}{\sigma_{h}^{2}} \end{bmatrix}_{m = 1}^{P_{\texttt{x}}}, \left(\diag(\widetilde{\vec{\eta}_{\texttt{x}}}) + \begin{bmatrix}\dfrac{\widetilde{\mat{G}_{\texttt{x},m}^{\top}\mat{G}_{\texttt{x},o}}}{\sigma_{h}^{2}} \end{bmatrix}_{m = 1, o = 1}^{P_{\texttt{x}}, P_{\texttt{x}}}\right)^{-1} \right) \\
	q(\mat{H}_{\texttt{x}}) &= \prod \limits_{i = 1}^{N_{\texttt{x}}} \mathcal{N}\left(\vec{h}_{\texttt{x},i}; \Sigma(\vec{h}_{\texttt{x},i}) \left(\sum \limits_{m = 1}^{P_{\texttt{x}}} \dfrac{\widetilde{e_{\texttt{x},m}} \widetilde{\vec{g}_{\texttt{x},m,i}}}{\sigma_{h}^{2}} + \dfrac{\widetilde{\mat{H}_{\texttt{z}}} (\vec{y}^{i})^{\top}}{\sigma_{y}^{2}} \right), \left(\dfrac{\mat{I}}{\sigma_{h}^{2}} + \dfrac{\widetilde{\mat{H}_{\texttt{z}} \mat{H}_{\texttt{z}}^{\top}}}{\sigma_{y}^{2}} \right)^{-1}\right).
\end{align*}

%%%%%%%%%%%%%%%%%%%%%%%%%%%%%%%%%%%%%%%%%%%%%%%%%%%%%%%%%%%%%%%%%%%%%%%%%%%%%%%%%%%%%%%%%%%%%%%%%%%%

\end{document}